\documentclass{article}
\usepackage[utf8]{inputenc}
\usepackage[T1]{fontenc}
\usepackage{todonotes}
\usepackage{placeins}
\usepackage{graphicx}
\usepackage{multirow}
\usepackage{caption}
\usepackage{subcaption}
\usepackage[a4paper]{geometry}
\usepackage{amssymb}
\usepackage{amsmath}
\usepackage{caption}
\usepackage{hyperref}
\usepackage{ctable}
\usepackage{bm}
\usepackage{authblk}
\usepackage[maxnames=50,sorting=nty]{biblatex}
\newcommand{\vect}[1]{\mathbf{#1}}

\newcommand{\R}{\mathbb{R}}
\newcommand{\set}[1]{\mathcal{#1}}

\definecolor{mypink1}{RGB}{255, 105, 180}

\title{Hyperspectral classification of blood-like substances using machine learning methods combined with genetic algorithms in transductive and inductive scenarios}

\author[1]{Filip Pałka}
\author[1]{Wojciech Książek}
\author[1,2,*]{Paweł Pławiak}
\author[2]{Michał Romaszewski}
\author[2]{Kamil Książek}
\affil[1]{Department of Computer Science, Faculty of Computer Science and Telecommunications, Cracow University of Technology, Krakow, Poland, filip.palka@pk.edu.pl, wojciech.ksiazek@pk.edu.pl, plawiak@pk.edu.pl}
\affil[2]{Institute of Theoretical and Applied Informatics, Polish Academy of Sciences, Gliwice, Poland, plawiak@iitis.pl, mromaszewski@iitis.pl, kksiazek@iitis.pl}
\affil[*]{plawiak@pk.edu.pl or plawiak@iitis.pl}

\date{October 2020}

\begin{document}

\maketitle

\begin{abstract}
This study is focused on applying genetic algorithms (GA) to model and band selection in hyperspectral image classification. We use a forensic-inspired data set of seven hyperspectral images with blood and five visually similar substances to test GA-optimised classifiers in two scenarios: when the training and test data come from the same image and when they come from different images, which is a more challenging task due to significant spectra differences. In our experiments we compare GA with a classic model optimisation through grid search. Our results show that GA-based model optimisation can reduce the number of bands and create an accurate classifier that outperforms the GS-based reference models, provided that during model optimisation it has access to examples similar to test data. We illustrate this with experiment highlighting the importance of a validation set.
\end{abstract}

\section{Introduction}
\label{sec:introduction}
Genetic optimisation, inspired by natural evolution, is a well-known heuristic optimisation and search procedure that can be used for both feature and model selection in Machine Learning. The focus of this paper is the use of Genetic Algorithms (GA) in order to train accurate hyperspectral classifiers. A hyperspectral classifier aims to assign pixels in a hyperspectral image to predefined classes e.g. different types of crops in an image of agricultural area. A hyperspectral pixel is a vector of measurements (typically, reflectance values) corresponding to specific band - a narrow wavelength range of the electromagnetic spectrum. Since materials in the imaged scene uniquely reflect, absorb, and emit electromagnetic radiation based on their molecular composition and texture, hyperspectral classification allows to accurately distinguish between them~\cite{ghamisi2017advanced}.

However, there are several challenges related to the task, such as the huge volume of images, their high dimensionality, redundancy of information in hyperspectral bands and the presence of noise introduced by acquisition process and calibration procedures~\cite{bioucas2013hyperspectral}. In addition, observed spectra are mixtures (e.g. linear combinations) of material spectra in the imaged scene~\cite{bioucas2012hyperspectral}.

One particular challenge lies in the availability and quality of training data i.e. selection of a training set. Typically, due to the high cost of generating hyperspectral training examples~\cite{landgrebe2005signal}, training sets in hyperspectral classification are small. However, when training pixels are randomly, uniformly sampled from the classified image itself, it is possible to achieve high accuracy even for very small training sets of 5-15 examples per class e.g. by exploiting the spatial-spectral structure of the image and using semi-supervised learning~\cite{romaszewski2016}. This is because hyperspectral images provide highly distinctive features and because classes are usually relatively large in the image. In such problems we may be more interested in finding the best assignment of pixels to classes than in the classification function itself. Therefore, referring to the concept of transductive learning proposed by Vapnik \cite{vapnik1977structural}, we call such scenario a \textit{Hyperspectral Transductive Classification} (HTC) problem. 

The challenge is elevated when training pixels come from a different image than test pixels. In such a case, differences in acquisition environment (e.g. light intensity, time differences) and in class spectra (e.g. different background materials in spectral mixtures) may be perceived as a complex noise. In such scenario the classifier is expected to generalise and compensate the differences between the training set and classified data. In contrast to the HTC scenario, which treats the image as a ,,closed world'', we call this scenario the \textit{Hyperspectral Inductive Classification} (HIC), emphasising the importance of finding the best classification function. The HIC scenario shares similarities with hyperspectral target detection problem~\cite{manolakis2003hyperspectral}, where spectra to be found in an image commonly come from spectral libraries.

Genetic Algorithms~\cite{holland1992,rutkowski2008}  are well-established techniques for selection of features and optimisation of classifier parameters. GA are based on natural selection, inheritance and the evolutionary principle of survival of the best adapted individuals. Their advantages compared to the classic feature and model selection procedures such as grid search are e.g. a) resistance to local extremes; b) the ability to control selective pressure (exploration and exploitation) from global to local search; c) ease of application due to feature selection being combined with parameter optimization. These advantages resulted in GAs being frequently used for hyperspectral band selection~\cite{ma2003application} and classification of multispectral~\cite{sukawattanavijit2017ga} and hyperspectral data~\cite{kumar2016dimensionality}.

However, in most of these works, GAs are applied for a problem corresponding to the HTC scenario, typically using well-known hyperspectral datasets such as the `Indian Pines' or the `University of Pavia' images. Under such conditions simultaneous optimisation of classifier parameters with band selection allows to achieve high classification accuracy~\cite{zhuo2008genetic}.

Our main goal is to test an compare the accuracy of GA-based classifiers in both the transductive and inductive hyperspectral classification scenarios. In our experiments we use a dataset described in~\cite{romaszewski2020dataset} that consists of of multiple hyperspectral images with blood and blood-like substances. The dataset is inspired by problems related to forensic analysis e.g. the detection of blood. However, our focus is on the problem of classification i.e. distinguishing between classes corresponding to visually similar blood-like substances in the images. We use multiple images with the same classes but with significant spectral differences, to compare the HTC and the HIC scenarios. We analyse the impact of GA on the classification accuracy in comparison to the grid-search parameter selection using multiple state-of-the-art hyperspectral classifiers. Our thesis is that hyperspectral classification with a GA applied to optimisation of classifier parameters and band selection allows to obtain more accurate classifiers than the grid search in both the HTC and HIC scenarios.

The paper is organised as follows. Section 2 provides a brief overview of the literature. Section 3 describes the materials used (dataset)  and methods (the stages of processing and analysis). The conducted experiment is described in section 4 and the results obtained are presented in section 5. Finally, sections 6 and 7 present conclusions and a summary.

\section{State of the art}
\label{sec:SOA}
\subsection{Hyperspectral classification}
In this paper we focus on spectral classification~\cite{ghamisi2017advanced} which uses only spectral vectors. The leading approaches involve the use of Support Vector Machines~\cite{melgani2004classification}, Extreme Learning Machines and their Kernel-based variants~\cite{pal2013kernel} or Multinomial Logistic Regression~\cite{khodadadzadeh2014subspace}. In order to further improve classification accuracy, spectral-spatial approaches~\cite{ghamisi2018new} which make use both pixel spectra and their spatial position in the image are employed. In particular, a combination of spatial-spectral and semi-supervised approaches allows to reach a high classification accuracy even for a small training set~\cite{romaszewski2016}.
Recently, deep learning methods~\cite{li2019deep} are popular, although their limiting factor is the fact that they usually require relatively large training sets. However, some works, such as e.g. the approach presented in~\cite{fang2018semi}, based on residual networks, seem to be able to significantly reduce this dependency.

\subsection{Evolutionary computation and genetic algorithms}

The advantages of techniques based on  computational intelligence \cite{Engelbrecht_2007} methods lie in the properties inherited from their biological counterparts: learning and generalization of knowledge (artificial neural networks \cite{Tadeusiewicz2015}), global optimization (evolutionary computation \cite{Back_1997}) and the use of imprecise terms (fuzzy systems \cite{Nguyen_1998}). 
The inspiration to undertake research on evolutionary computation (EC) \cite{Back_1997} was the imitation of nature in its mechanism of natural selection, inheritance and functioning. 
Genetic algorithms (GA)~\cite{Sivanandam_2008} are a part of evolutionary computation techniques, used with success in the field such as vehicle routing problem \cite{Park_2021}, feature selection \cite{Zhou_2021}, optimization \cite{DAngelo_2021}, heart sound segmentation \cite{Alonso_2021} or traveling salesmen problem \cite{Dong_2021}

Genetic algorithms are one of the leading approaches to solve optimisation problems~\cite{rutkowski2008}. Optimization problems are computationally complex,therefore they are often solved with heuristic methods, which make it possible to find a near-optimal solution faster. Genetic algorithm works by creating a population consisting of a selected number of individuals, each of them representing one solution to the problem. Then, from among all the individuals, those with the best results are selected, and then subjected to genetic operators, they create a new population. In particular, this technique can be applied for model selection, to find parameters of a machine learning model and simultaneously perform feature selection as e.g. in works heart arrhythmia detection \cite{plawiak2020arrhytmia}, \cite{plawiak2018myocardium}, early diagnosis of hepatocellular cancer \cite{ksiazek2019}, or prediction of credit scoring \cite{plawiak2020scoring}.

\subsection{Hyperspectral classification and band selection with GA}
\label{sec:soa-gahsi}
GAs have been used many times for classification and selection of characteristic wavelengths in hyperspectral data. For example, in~\cite{ma2003application} authors use GA to find small subsets of the most distinctive bands. In~\cite{kumar2016dimensionality} GAs are applied for band selection in prepossessed hyperspectral images in order to classify them. In~\cite{pedergnana2013novel} GA optimization is used to divide hyperspectral bands into three classes related to their discriminative power in the classification task. Authors verify their results using three standard hyperspectral datasets i.e. the `University of Pavia', `Indian Pines' and `Hekla'. The use of GA for simultaneous optimization of SVM parameters and band selection in HSI classification is presented in~\cite{zhuo2008genetic}. A similar scheme for multispectral data is used in~\cite{sukawattanavijit2017ga} -- authors emphasize the advantage of GA algorithms over parameter optimization using grid search. A very interesting use of GA is presented in~\cite{nagasubramanian2018hyperspectral}: authors apply GA to large number of hyperspectral cubes (111 images) in order to determine a subset of wavelengths characteristic for identification of charcoal rot disease in soybean stems.

\section {Materials and methods}
\begin{figure*}[ht!]
	\centering
	\begin{subfigure}[b]{0.49\textwidth}
		\includegraphics[width=1.0\linewidth]{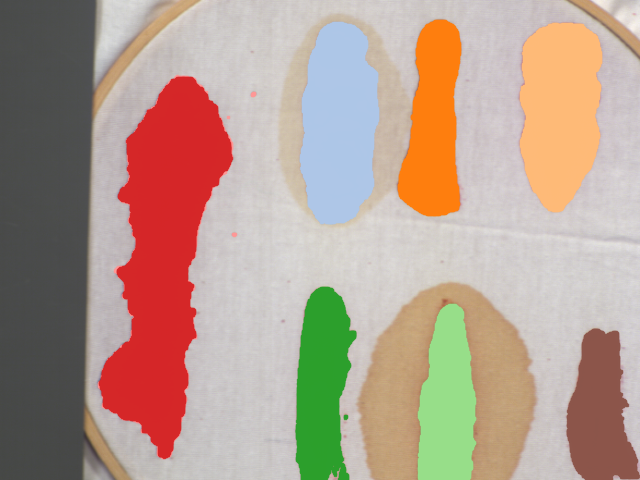}
		\caption{Class GT for image \textit{F(1)}}
	\end{subfigure}
	\begin{subfigure}[b]{0.49\textwidth}
		\includegraphics[width=1.0\linewidth]{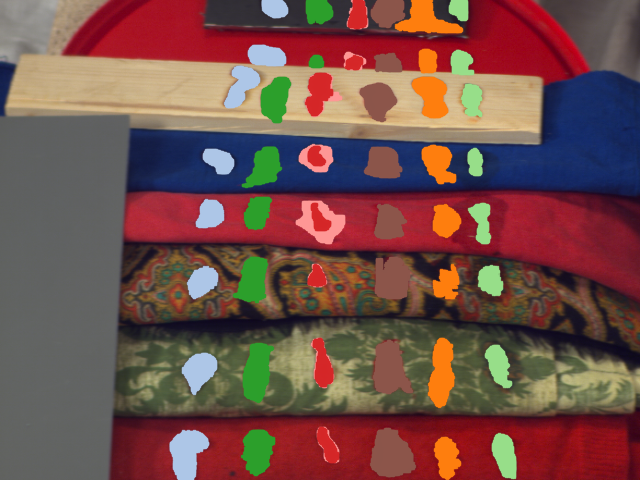}
		\caption{Class GT for image \textit{E(1)}}
	\end{subfigure}
	\begin{subfigure}[b]{0.49\linewidth}
		\includegraphics[width=1.0\linewidth]{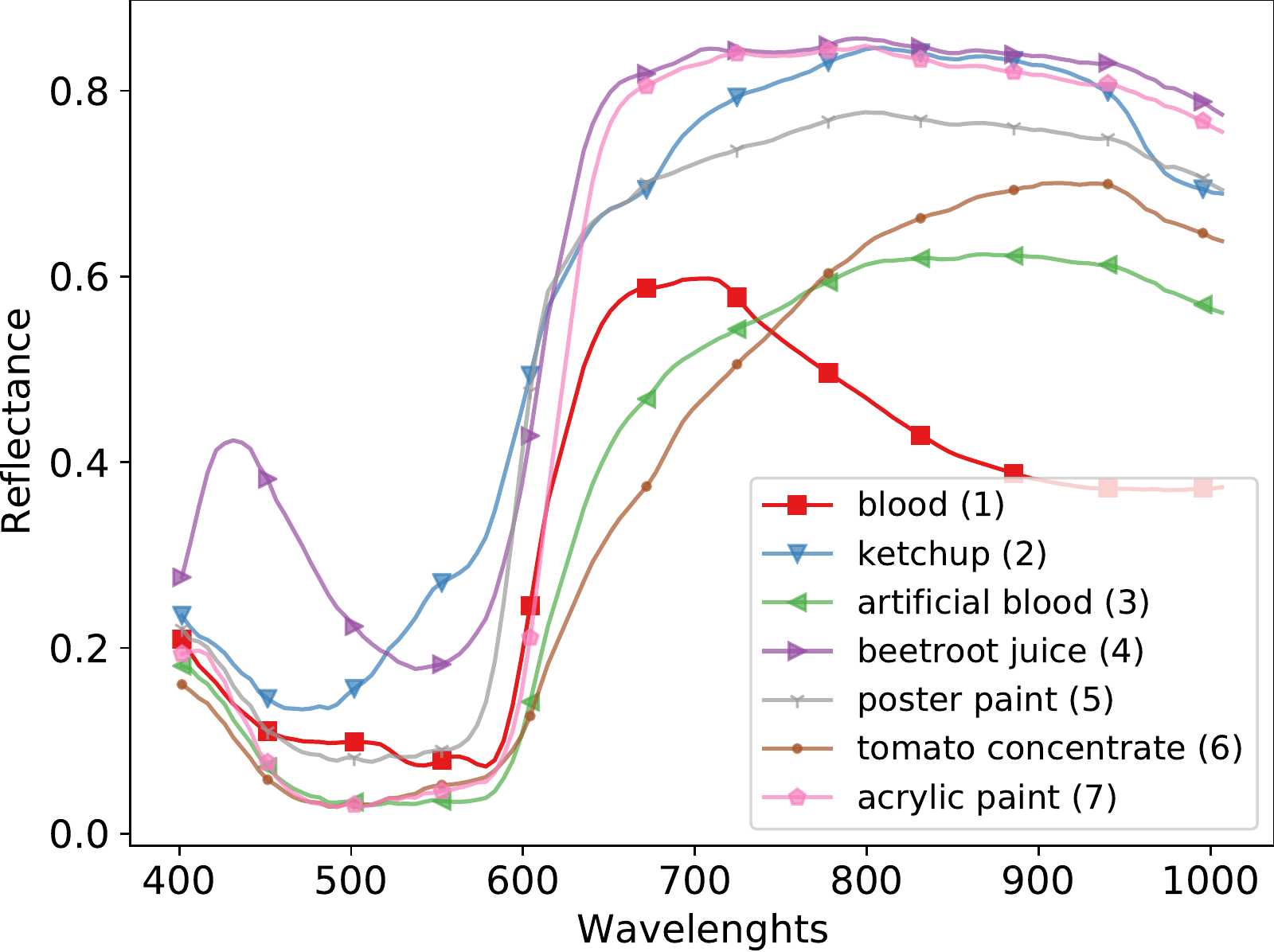}
		\caption{Class spectra in the image \textit{F(1)}}
	\end{subfigure}
	\begin{subfigure}[b]{0.49\linewidth}
	\includegraphics[width=1.0\linewidth]{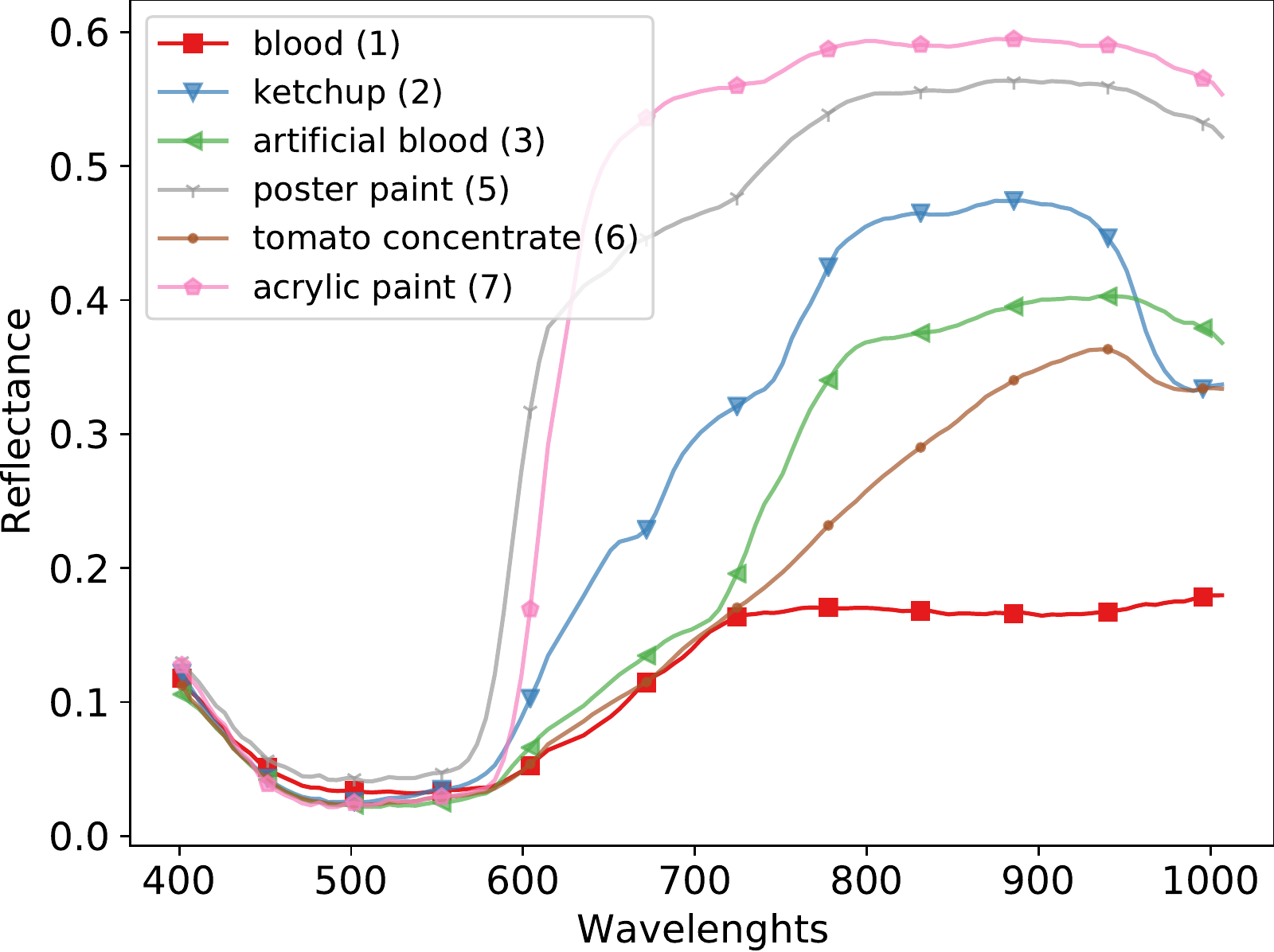}
	\caption{Class spectra in the image \textit{E(1)}}
	\end{subfigure}
	\begin{subfigure}[b]{0.49\linewidth}
		\includegraphics[width=1.0\linewidth]{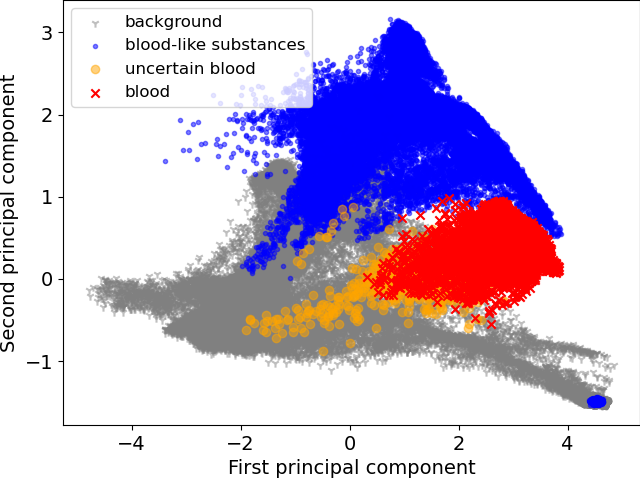}
		\caption{PCA visualisation for the image~\textit{F(1)}}
	\end{subfigure}
	\begin{subfigure}[b]{0.49\linewidth}
	\includegraphics[width=1.0\linewidth]{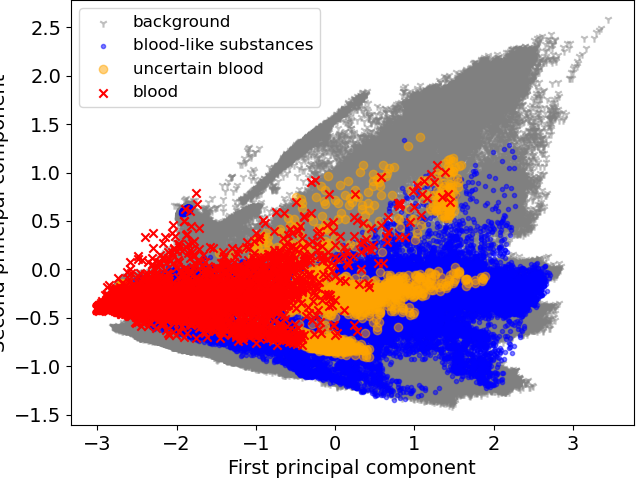}
	\caption{PCA visualisation for the image~\textit{E(1)}}
	\end{subfigure}
\caption{Dataset visualisation. Upper panels present classes as a coloured ground truth on RGB images created from hyperspectral cubes. Middle panels present mean class spectra. Bottom panels present PCA projection of data on first two principal components. Images come from~\cite{romaszewski2020dataset}.}
	\label{fig:dataset}
\end{figure*}

\subsection{Dataset}
\label{sec:dataset}
We used the dataset described in~\cite{romaszewski2020dataset}, consisting of multiple hyperspectral images of blood and blood-like substances such as artificial blood, tomato concentrate or poster paint. Hyperspectral pixels where these substances are visible were annotated by authors.

Images in the dataset were captured using SOC710 hyperspectral camera operating in spectral range 377-1046 nm with 128 bands. As suggested in~\cite{romaszewski2020dataset} noisy bands [0-4], [48-50] [122-128] were removed which left 113 bands. Two types of images are used in our experiments: the ,,\textit{Frame}'' images, denoted as~\textit{F} in~\cite{romaszewski2020dataset},  present classes on a uniform, white background; the ,,\textit{Comparison}'' images (denoted as \textit{E}) present classes on diverse backgrounds consisting of multiple materials and fabrics.

We used images captured in days $\{1,7,21\}$. Following the convention from~\cite{romaszewski2020dataset}, we denoted the day of acquisition after the scene name, in brackets e.g. \textit{F(1)} for the scene ,,\textit{Frame}'' from day 1.
Visualisation the dataset is presented in Fig.~\ref{fig:dataset}.

\subsection{Data preprocessing}
\label{sec:preprocessing}
The aim of the initial preprocessing applied to dataset images was to reduce noise and compensate for uneven lighting. The following sequence of transformations was applied to every image: 

\begin{enumerate}
\item \textbf{Median filter}
Images were smoothed with a spatial median filter with a window size of one pixel. This operation is intended to reduce the noise in spectra, using the fact that classes are to be significantly larger than a single pixel.

\item \textbf{Spectra normalization}
As suggested in~\cite{romaszewski2020dataset}, the spectrum of each pixel was divided by its median. The purpose of this normalisation is to compensate for uneven lighting in the image.

\item \textbf{Removal of noisy bands}
As suggested in~\cite{romaszewski2020dataset}, noisy bands [0-4], [48-50] and [122-128] were removed, leaving 113 bands.

\end{enumerate}

\subsection{Feature extraction}
\label{sec:extraction}
\begin{figure}[h!]
	\centering
	\begin{subfigure}[b]{0.32\textwidth}
		\includegraphics[width=1.0\linewidth]{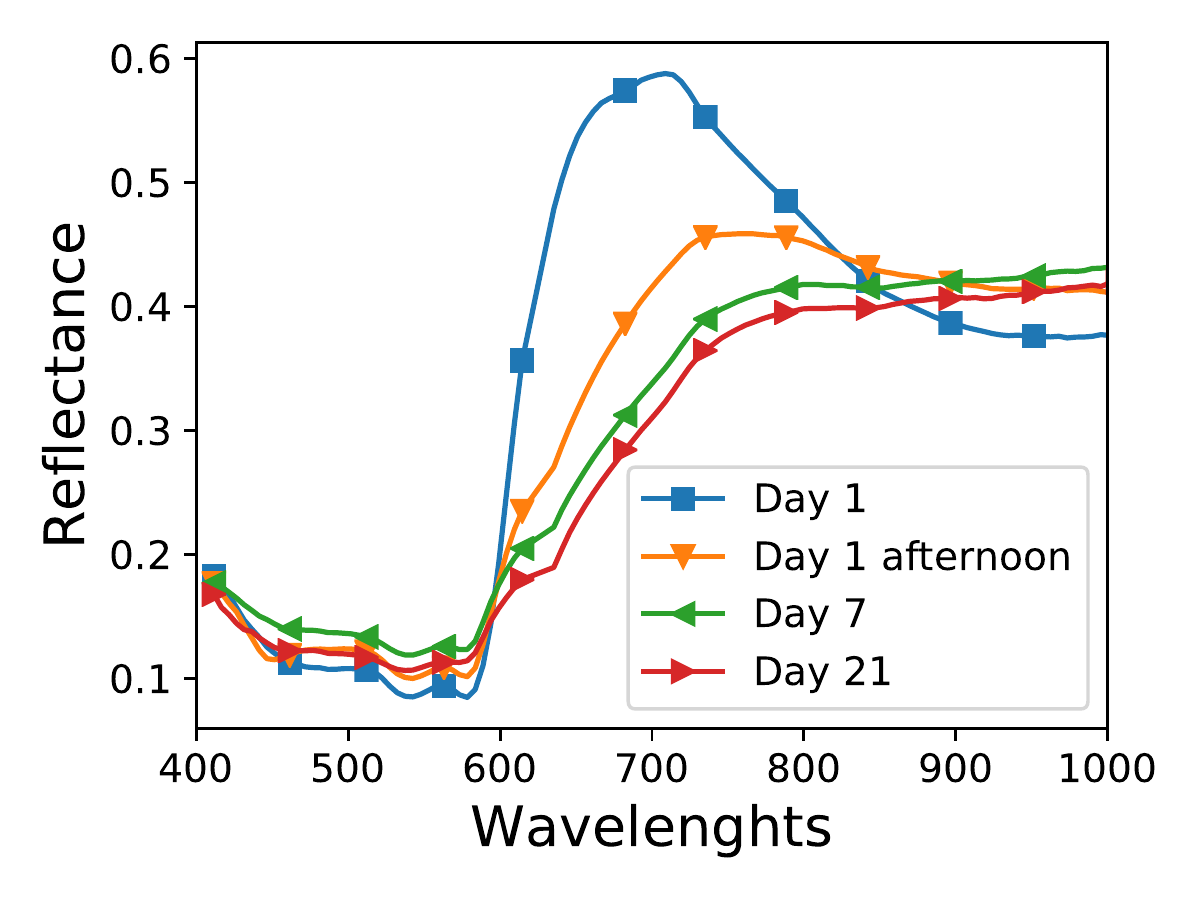}
		\caption{Raw spectra}
	\end{subfigure}
	\begin{subfigure}[b]{0.32\textwidth}
		\includegraphics[width=1.0\linewidth]{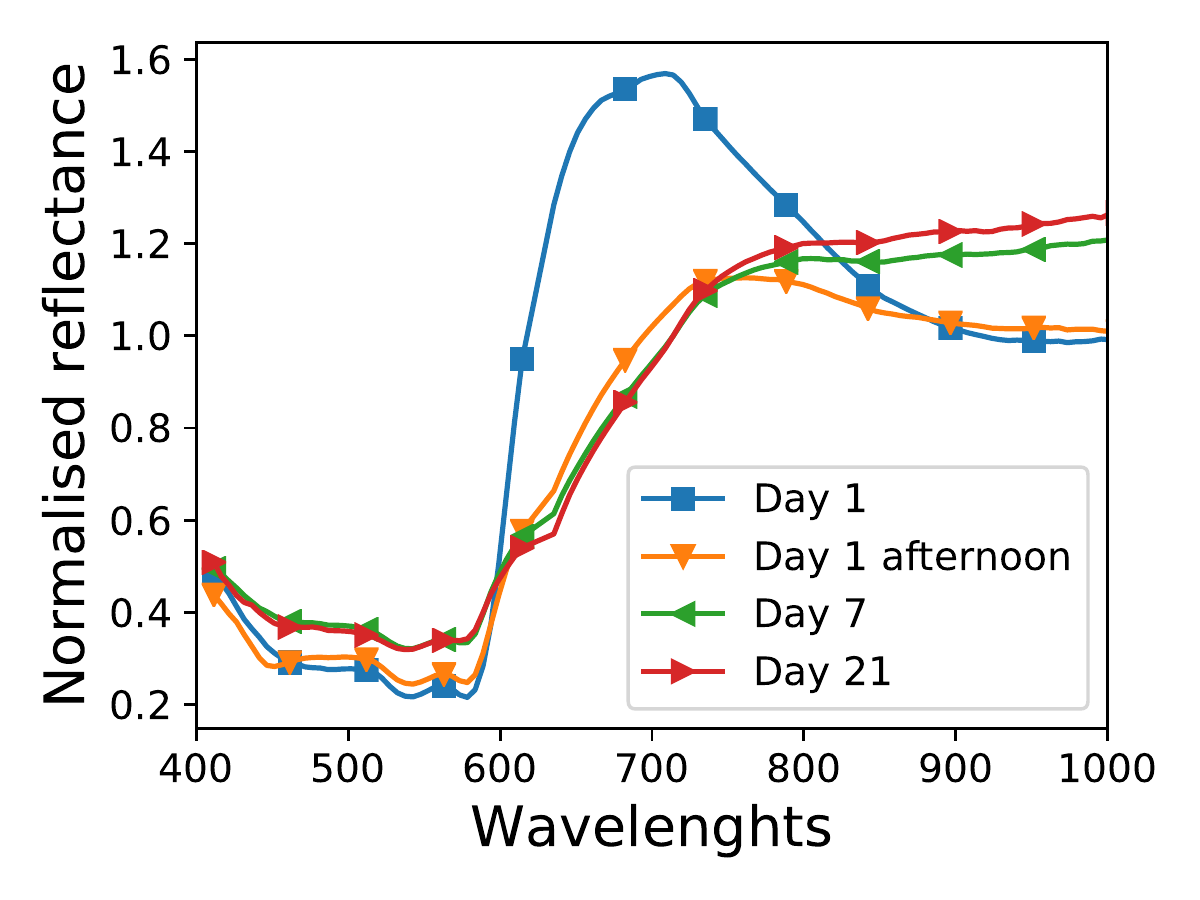}
		\caption{Preprocessed spectra}
	\end{subfigure}
	\begin{subfigure}[b]{0.32\linewidth}
		\includegraphics[width=1.0\linewidth]{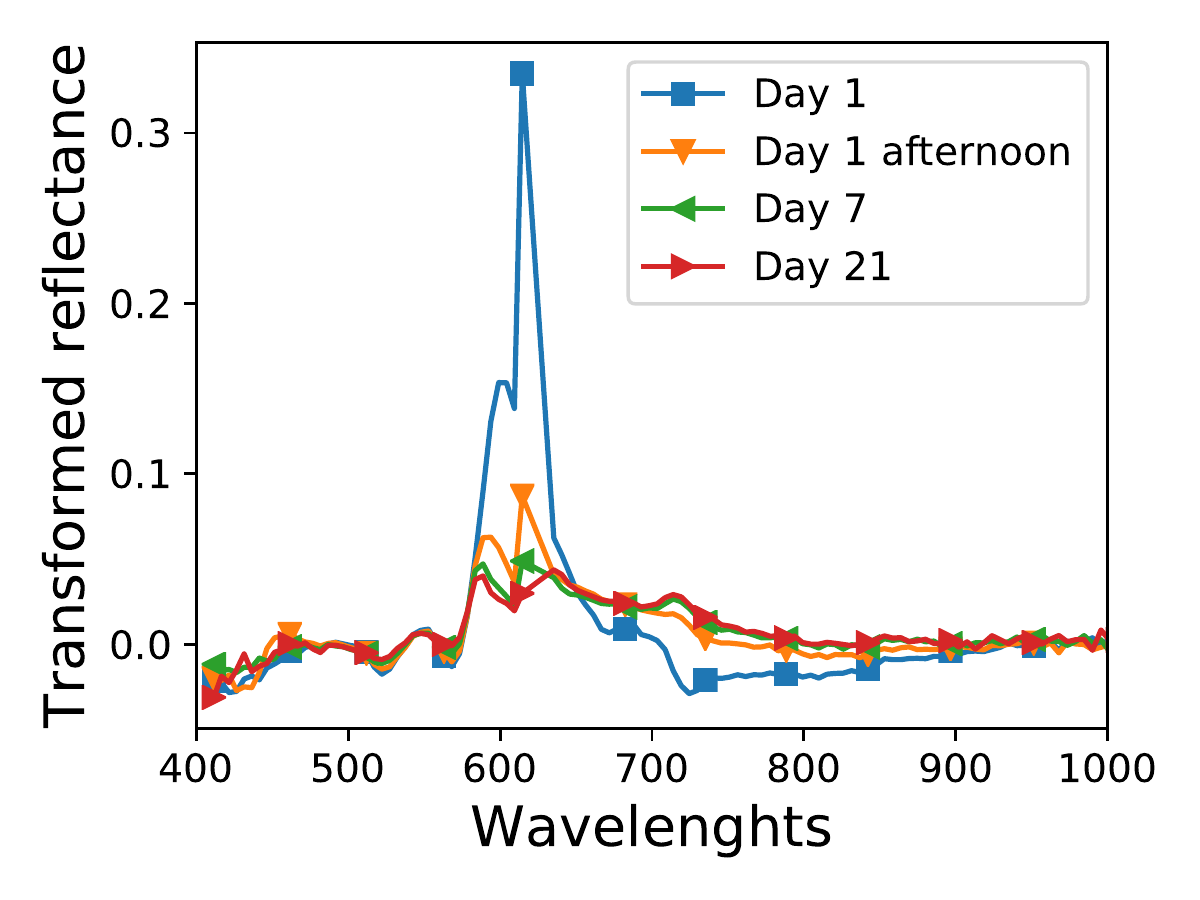}
		\caption{Transformed spectra}
	\end{subfigure}
    \caption{Visualisation of the impact of preprocessing on example spectra of the `blood' class from the dataset.}
	\label{fig:blood_transformations}
\end{figure}
In our experiments we used derivative transformation to highlight important features of  spectra. Derivative analysis~\cite{tsai1998derivative} is a well-known method for transforming spectral signatures. Derivatives are sensitive to the shape of spectra, therefore they are particularly effective in differentiating signals with characteristic spectral responses, such as e.g. hemoglobin response in blood~\cite{majda2018hyperspectral}, visible as peaks in wavelengths $\sim542$ nm and $\sim576$ nm (called $\alpha$ and $\beta$ bands). We used first order derivatives, computed as a difference between neighbouring bands.

A visualisation of the impact of preprocessing and feature extraction on example spectra is presented in Fig~\ref{fig:blood_transformations}.

\subsection{Classification algorithms}
\label{sec:svm}
\subsubsection{Support Vector Machines}
In this work we focus on a Support Vector Machine~\cite{scholkopf2018learning} (SVM) classifier, that is accurate in hyperspectral classification problems~\cite{ghamisi2017advanced}, including classification of hyperspectral forensic data~\cite{glomb2018application} and is well suited for optimisation with GA~\cite{zhuo2008genetic}. HSI classification with SVM can be described as follows:

Given a training set of labelled examples
\begin{equation}
\set{T}=\left\{(\mathbf{x}_i,y_i),i=1,\dots,n\right\}
\quad
\mathbf{x}_i\in\set{X}
\quad
y_i\in\set{Y},
\end{equation}
where $\set{X}$ denotes a set of examples (e.g. hyperspectral pixels) and $\mathcal{Y}=\{-1,1\}$ denotes the set of labels, the SVM classifies a hyperspectral example $\mathbf{x}\in\set{X}\subset\R^{d}$ using a function:
\begin{equation}
f(\mathbf{x}) = \mathrm{sgn}\Big(\sum_{i=1}^{n}y_i\beta_i K(\mathbf{x},\mathbf{x}_i)+b\Big),
\end{equation}
where $\beta_i \geq 0$ and $b$ are coefficients computed through Lagrangian optimization (margin maximization on the training set). The kernel function $K:\set{X}\times\set{X}\rightarrow\R$ is used to compute the similarity measure between the classified example $\mathbf{x}$ and every training instance $\mathbf{x}_i$. 

We use three kernel functions: 
\begin{itemize}
    \item Gaussian radial basis function (RBF) $K(\mathbf{x}_i,\mathbf{x}_j)=\exp(-\gamma||\mathbf{x}_i-\mathbf{x}_j||^2)$, parameterised with $\lbrace\gamma\rbrace$,
    \item sigmoid kernel $K(\mathbf{x}_i,\mathbf{x}_j)=\text{tanh}(\gamma\vect{x}_i^T\vect{x}_j+c_0)$ parametrised with $\lbrace\gamma, c_0\rbrace$
    \item polynomial kernel $K(\mathbf{x}_i,\mathbf{x}_j)=(\gamma \vect{x}_i^T\vect{x}_j+c_o)^d$ parametrised with $\lbrace\gamma, c_0, d\rbrace$ that can be simplified to the linear kernel $K(\mathbf{x}_i,\mathbf{x}_j)= \vect{x}_i^T\vect{x}_j$ when parameters $d=c_0=0$.
\end{itemize}

In addition to parameters of a chosen kernel, the SVM has an additional regularisation parameter $C$, that controls the balance between maximisation of margin between classes and missclassification of examples. The value of this parameter must be fitted to a given problem, typically through cross-validation. However, the use of GA for selecting parameters is complicated by the fact that the value of $C$ is unbounded from above. Therefore, in our experiments we used a classifier proposed in~\cite{scholkopf2000new}, namely the $\nu$-SVM that uses a bounded regularisation parameter $\nu\in(0,1\rangle$, which is an upper bound on the fraction of missclasified examples from the training set and a lower bound on the fraction of support vectors.

\subsubsection{K-Nearest Neighbor (KNN)}
The K-nearest neighbors algorithm (KNN)~\cite{knn2016} belongs to the family of non-parametric models. The principle of operation of the algorithm is based on making predictions based on the closest neighborhood of an example. A new, unclassified sample is labelled through a majority vote of a neighborhood of a fixed size weighted by the distance of this sample from the voting neighbors. In our experiments we used the Euclidean, the Manhattan and the Chebyshev distance measures.

\subsubsection{Multilayer Perceptron}
A Multilayer Perceptron \cite{mlp2016} is a neural network composed of a combination of individual perceptrons that together form a multilayer structure. The most frequently distinguished layers are the input, hidden and output layer. Each layer may have a different number of neurons. Advanced network models consist of multiple hidden layers. MLP is typically trained using a backpropagation algorithm. Despite its simplicity, the MLP achieves high accuracy on hyperpsectral data and is often used as a reference method for other algorithms~\cite{ghamisi2017advanced}.

\subsection{Model selection with Genetic Algorithms}
\ctable[
cap     = Chromosome structure,
caption = The structure of a chromosome corresponding to optimized parameters of the $nu$-SVM classifier along with selected hyperspectral bands,
label   = tab:geneticAlgorithmParameters,
pos     = ht]
{ll}{\tnote[a]{Kernel function}\tnote[b]{Parameter of the polynomial kenrel}
\tnote[c]{Parameter of the RBF kenrel}
\tnote[d]{Parameter of the polynomial and sigmoid kenrel}}{\FL
Parameter&Range of values\ML
$K$\tmark[a]&$\{$RBF, polynomial, sigmoid$\}$\NN
$\nu$&$\langle0.001,0.4\rangle$\NN
$d$\tmark[b]&$\langle1,5\rangle$\NN
$\gamma$\tmark[c]&$\langle0.001,5\rangle$\NN
$c_0$\tmark[d]&$\langle0.01,10\rangle$\NN
band $1\ldots113$ &$\{\text{selected},\text{not selected}\}$\LL
}

\begin{figure}[ht]
    \centering
    \includegraphics[scale=0.7]{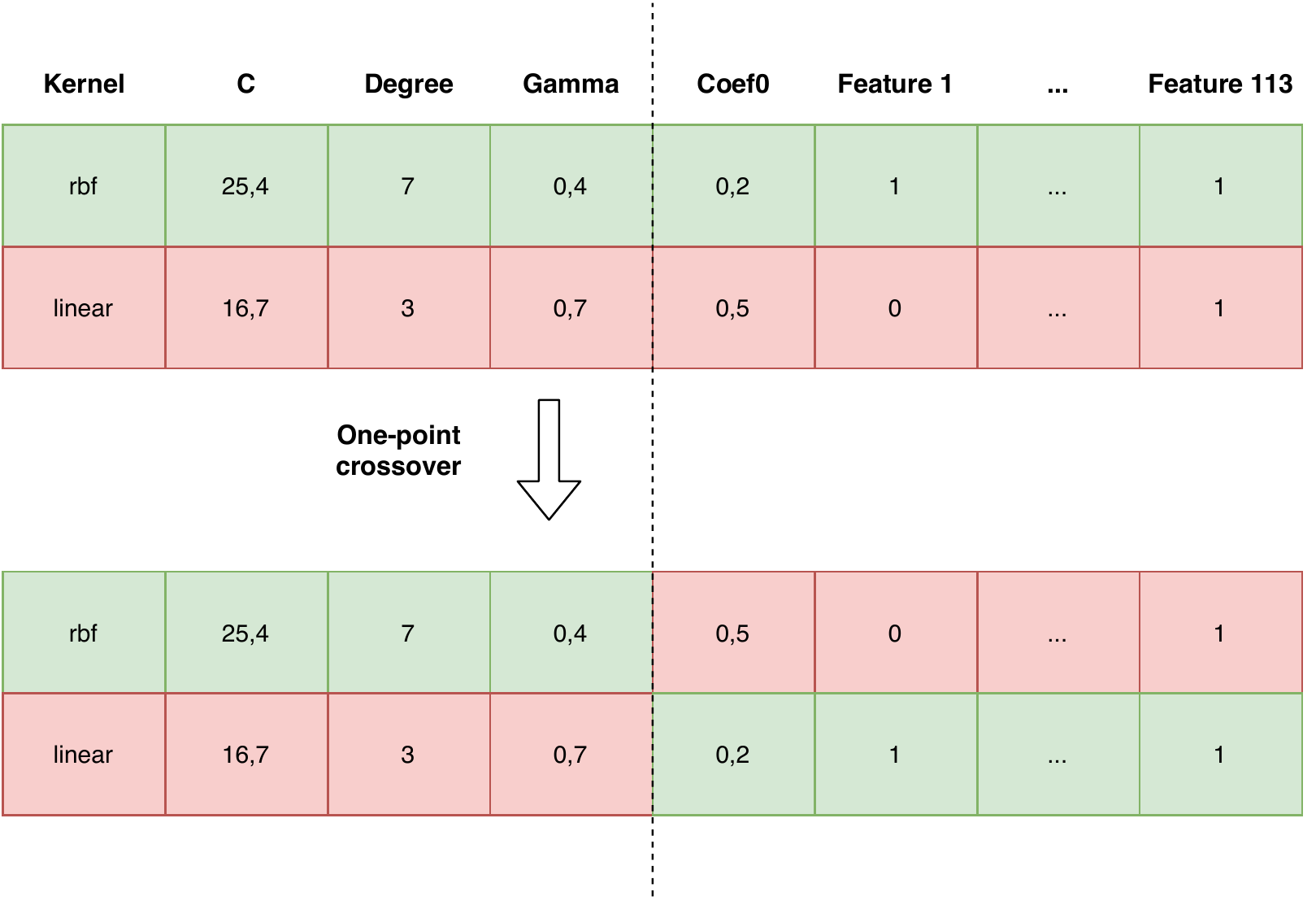}
    \caption{Visualisation of a one-point crossover between two individuals.}
    \label{fig:ag_chromosome_crossover}
\end{figure}

We use the genetic optimisation~\cite{rutkowski2008} to simultaneously select parameters of a machine learning model and perform feature selection. The $\nu$-SVM~\cite{scholkopf2018learning} classifier was chosen for this type of optimization due to its bounded parameterization of the margin (see Sec.~\ref{sec:svm}).

Taking advantage of the GA capabilities, which allow for the optimization of many parameters at once, in our implementation, the type of kernel function, kernel parameters, the regularization parameter and feature (hyperspectral band) selection are performed simultaneously.
Table~\ref{tab:geneticAlgorithmParameters} presents the structure of a single individual. In our implementation, this individual consists of one chromosome. 
The chromosome consists of 5 genes  responsible for kernel type and its parameters, and 113 genes responsible for hyperspectral bands.

Figure \ref{fig:ag_chromosome_crossover} shows an example crossover between two individuals (i.e. classifiers). We observed that high probabilities of crossing and mutation have a positive effect on the search space i.e. it allows to better search the search space and check more solutions, which reduces the chances of finding the locally optimal solution~\cite{grefenstette1992genetic}.
Thanks to the elitist strategy there is a certainty that the best individual found will not be lost. The mutation of an individual consists in the modification of a single gene in the chromosome. If it is a gene responsible for a parameter of the support vector machine, its value is replaced by the new value of the given parameter from the set range (acceptable values are shown in table \ref{tab:geneticAlgorithmParameters}). If we draw a gene that represents a feature, its value is replaced by the opposite one e.g from `not selected' (0) to `selected' (1). 
Values of our genetic algorithm parameters are presented in Tab.~\ref{tab:ga-parameters}. 

\ctable[
cap     = GA parameters,
caption = Parameters of the GA used in experiments,
label   = tab:ga-parameters,
pos     = ht]
{ll}{\tnote{own implementation}}{\FL
Parameter&Value\ML
Size of the population&200\NN
Number of epochs&100\NN
Fitness function&Accuracy\NN
Selection algorithm&Tournament selection, size 3\NN
Crossover method&Uniform crossover\NN
Mutation method&One-point mutation\tmark[1]\NN
Probability of crossover&0.8\NN
Probability of mutation&0.8\NN
Elitist strategy& 1 individual\LL     
}

\subsubsection{Model selection with grid search}
In our experiments, grid search (GS) was used as a reference method for model selection. 
In many works the SVM with the regularisation parameters $C$ (denoted SVC) with RBF kernel function is used as a reference algorithm, therefore we use it in addition to the $\nu-$SVM. We also test the KNN and MLP classifiers, described in Sec.~\ref{sec:svm}. Parameters of model selection with the GS are provided in~Tab.~\ref{tab:grid-search-parameters}.

\ctable[
cap     = GS parameters,
caption = Grid search (GS) parameters used in experiments,
label   = tab:grid-search-parameters,
pos     = ht]
{lll}{\tnote[a]{Kernel function}\tnote[b]{Parameter of the polynomial kenrel}
\tnote[c]{Parameter of the RBF kenrel}
\tnote[d]{Parameter of the polynomial and sigmoid kenrel}
\tnote[e]{Linear SVM, implemented in \emph{liblinear} library}}{\FL
Classifier&Parameter&Values\ML
\multirow{5}{*}{SVM}&$K$\tmark[a]&$\{$RBF, polynomial, sigmoid$\}$\NN
&$C$&$\langle0.001,1000\rangle$\NN
&$d$\tmark[b]&$\langle1,5\rangle$\NN
&$\gamma$\tmark[c]&$\langle0.001,5\rangle$\NN
&$c_0$\tmark[d]&$\langle0.01,10\rangle$\ML
\multirow{2}{*}{LSVM}\tmark[e]&loss&$\{$hinge, squared$\}$\NN
&$C$&$\langle0.001,1000\rangle$\ML
\multirow{5}{*}{$\nu-$SVM}&$K$\tmark[a]&$\{$RBF, polynomial, sigmoid$\}$\NN
&$\nu$&$\langle0.001,0.4\rangle$\NN
&$d$\tmark[b]&$\langle1,5\rangle$\NN
&$\gamma$\tmark[c]&$\langle0.001,5\rangle$\NN
&$c_0$\tmark[d]&$\langle0.01,10\rangle$\ML
\multirow{3}{*}{KNN}&dist. metric&$\{$Euclidean, Manhattan, Chebyshev$\}$\NN
&weights&$\{$uniform, distance$\}$\NN
&n. neighbors&$\langle1,20\rangle$\ML
\multirow{6}{*}{MLP}&no. hidden layers&$\langle1,3\rangle$\NN
&number of neurons&$\big\{\{1000\},\{30,30\},\{1000,1000\},\{1000,1000,1000\},$\NN 
&on consecutive layers&$\{1000,1000,1000\}\big\}$\NN
& dropout                &$\{0,0.5\}$\NN
& learning rate          &$\{0.1, 0.01, 0.001\}$\NN
& batch size             &$\{50, 100\}$\NN
& number of iteration    &$\{50, 100, 150,\ldots, 500\}$\LL
}

\FloatBarrier

\subsubsection{Implementation}
All experiments were implemented in python, using the scikit-learn~\cite{scikit-learn}, PyTorch~\cite{NEURIPS2019_9015} and DEAP~\cite{DEAP} libraries.

\subsubsection{Model performance metric}
Because the number of examples in classes in our data set is similar, we used the accuracy as a performance metric, defined as follows:

\begin{equation}
    accuracy = \frac{1}{N} \left(\sum_{i=1}^{N}\frac{TP+TN}{TP+FP+TN+FN}\right) \times 100\%,
\end{equation}
where $N$ is number of folds in cross validation, $TP$ denotes True Positives, $TN$ denotes True Negatives, $FP$ denotes False Positives and $FN$ denotes False Negatives.

\section{Experiments}
\label{sec:experiments}
The main idea behind our experiments is to perform model and feature selection with GA and compare these results with a diverse set of classifiers, trained classically i.e. with a grid-search. Referring to classification scenarios introduced in Sec.~\ref{sec:introduction} we consider three experimental scenarios:
\begin{enumerate}
    \item Hyperspecreal Transductive Classification (HTC) - training and test examples are randomly, uniformly selected from a single hyperspectral image.  
    \item Hyperspecreal Inductive Classification (HIC) - training and test examples are selected from different images. Typically training examples come from \emph{Frame} images and testing examples come from the \emph{Comparison} images. 
    \item Hyperspecreal Inductive Classification with a Validation Set (HICVS) - this scenario is similar to the HIC scenario: training examples come from \emph{Frame} images and testing examples come from the \emph{Comparison} images. However, model selection is performed using a separate validation set that is randomly, uniformly sampled from the \emph{Comparison} scene. This scenario is designed to test the capabilities of GA optimization under different conditions than in the HIC scenario, which will be discussed in detail in Sec.~\ref{sec:discussion}.
\end{enumerate}

\subsection{The scheme of experiments}
\label{sec:experimens_scheme}
\begin{figure}[ht]
    \centering
    \includegraphics[scale=0.6]{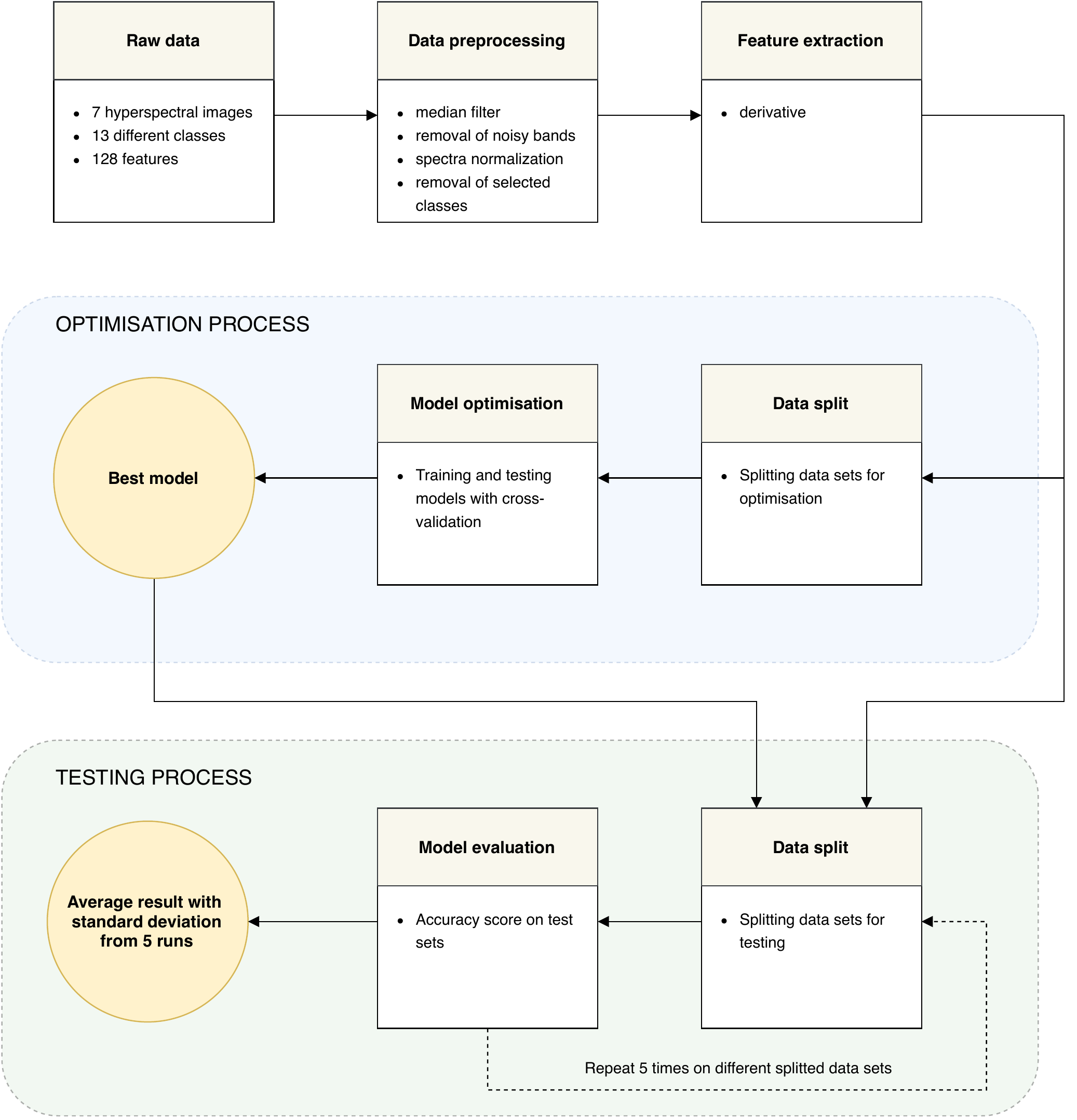}
    \caption{The overview scheme of experiments.}
    \label{fig:experiments_diagram}
\end{figure}
An overview schema of our experiments is presented in Fig.~\ref{fig:experiments_diagram}. The experiments can be divided into six stages:

\begin{enumerate}
    \item Raw data -- The data set consists of seven hyperspectral images from the data set described in Sec.~\ref{sec:dataset}. Every image has 128 hyperspectral bands. The images represent two scenes - the \emph{''Frame''} scene and the \emph{''Comparison''} scene. Four of the seven images show the \emph{''Frame''} scene, captured in days $\{1,1_a,7,21\}$, where the value $1_a$ represents the afternoon of the first day. The three \emph{''Comparison''} images were captured in days $\{1,7,21\}$. 
    \item Data preprocessing -- Data was transformed in accordance with methodology described in Sec.~\ref{sec:preprocessing}: in order to reduce the effect of noise and uneven lighting spectra were smoothed with the median window, normalised and noisy bands were removed. Background (unannotated pixels) and pixels from the class `beetroot juice' (class $4$) that is not present in all images were removed. Finally, the problem is posed as a six-class classification with classes $\set{Y}=\{1,2,3,5,6,7\}$.
    \item Feature extraction -- we used derivative transformation, described in Sec.~\ref{sec:extraction}.
    \item  Data split -- Data was divided into training and test sets. A detailed description of this stage is included in sections Sec.~\ref{sec:exp:HTC}, Sec.~\ref{sec:exp:HIC} and Sec.~\ref{sec:exp:HICVS}.
    \item Model optimization -- model selection was performed in o detailed description of this method in the context of the problem of classifying hyperspectral data using the support vector machine is described in section 3.3.4. The reference method used to compare results with the evolutionary algorithm was grid search. In both optimisation cases, accuracy was chosen as the evaluation method. During optimisation all models were trained and tested using cross-validation. The settings and details of the cross-validation varied depending on the scenario of the experiment. A detailed description of the optimisation process can be found in the description of the individual scenarios.
    \item Model evaluation - The final measure of classification evaluation is accuracy. After finding the best model in stage V, such a model is trained on the entire training set without cross-validation. Then it is tested on test sets. The test sets are created from both scenes: \emph{''Frame''} and the \emph{''Comparison''} scene. For reliable results, the training and testing process is repeated five times. The results from each repetition are saved and then the average with the standard deviation is calculated. The main measure of model evaluation is the classification accuracy on test scene.
\end{enumerate}

\subsection{Hyperspectral Transductive Classification (HTC)}
\label{sec:exp:HTC}
In the HTC training pixels are randomly, uniformly sampled from the same images as test pixels. This scenario bears resemblance to a common hyperspectral classification setting, when classifiers are tested e.g. using the `Indian Pines' data set~\cite{ghamisi2017advanced}. The aim of this experiment is to test the capability of classifiers to model classes and distinguish between them.

The training set is a combination of examples from all images i.e. \emph{''Frame''} and \emph{''Comparison''} scenes from all days. The training set consists of an equal number of examples from each class and each day. We used the size of the least numerous class among all the images (989) therefore, the training set consisted of 41,538 hyperspectral pixels (989 pixels * 6 classes * 7 images). 

After selecting the best parameters and features using cross-validation on the training set, classifiers were trained on the whole training set and tested on the remaining examples.

\subsection{Hyperspectral Inductive Classification (HIC)}
\label{sec:exp:HIC}
\begin{figure}[ht]
    \centering
    \includegraphics[scale=0.6]{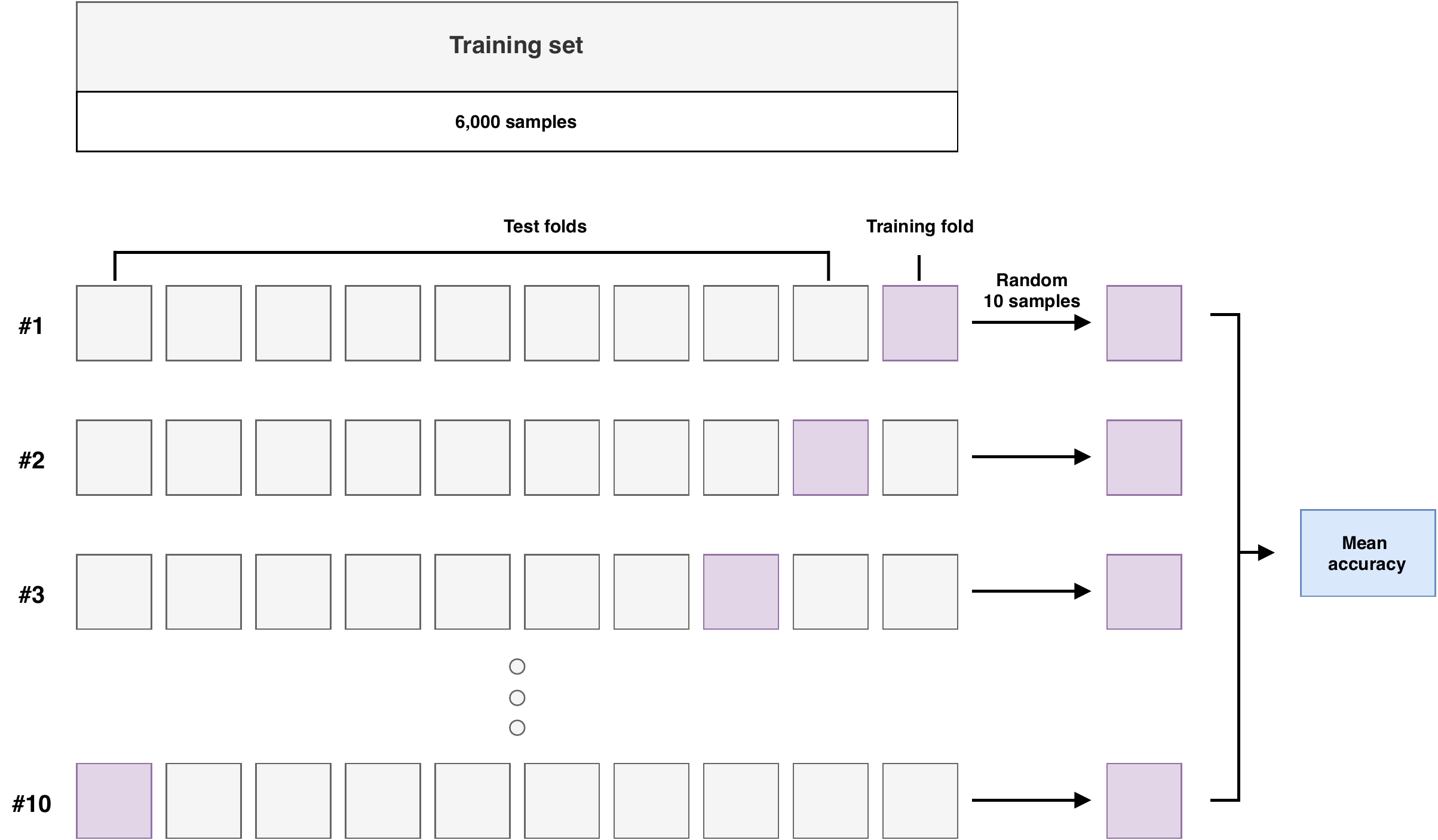}
    \caption{Visualisation of the model optimisation stage in the HIC scenario, using 10-fold cross-validation on a selected training set from \emph{''Frame''} images.}
    \label{fig:cross-validation-inductive}
\end{figure}
In the HIC scenario classifiers are trained on \emph{''Frame''} images and tested on \emph{''Comparison''} images. This scenario simulates a potential forensic application, where the model is prepared using laboratory samples and applied in the field in an unknown environment.

The training set consisted of examples sampled only from   \emph{''Frame''} images. The training set size was 6000 examples (250 examples from each class, from four available images). The test set consisted of a total of 82,097 examples from \emph{''Comparison''} scenes.

Each model was optimized in the process of a 10-fold cross-validation: every time one fold was used for training and the remaining ones for testing. Additionally, only a subset of $10$ randomly selected examples from each class in the training set are used for training in a single cross-validation iteration.
After the optimization stage, the best model was trained on examples in the training set and tested on a test set.

Visualisation of the model optimisation process in the HIC scenario using cross-validation is presented in Fig.~\ref{fig:cross-validation-inductive}. 

\subsection{Hyperspectral Inductive Classification with a validation set (HICVS)}
\label{sec:exp:HICVS}
\begin{figure}[ht]
    \centering
    \includegraphics[scale=0.65]{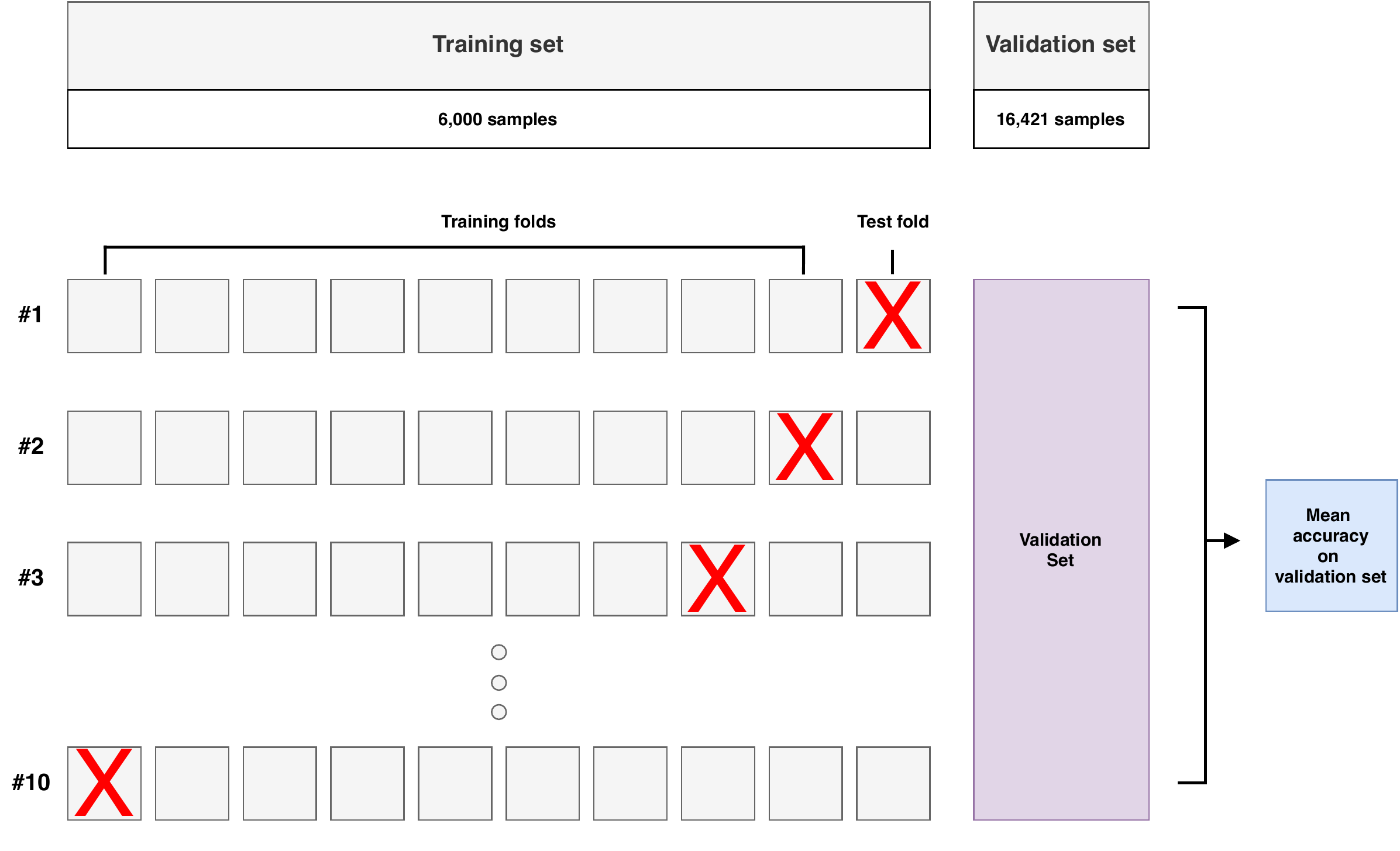}
    \caption{Visualisation of the model optimisation stage in the HICVS experiment with a small training set and 10-fold cross-validation}
    \label{fig:cross_validation_inductive__wv_smaller_set}
\end{figure}
\begin{figure}[ht]
    \centering
    \includegraphics[scale=0.65]{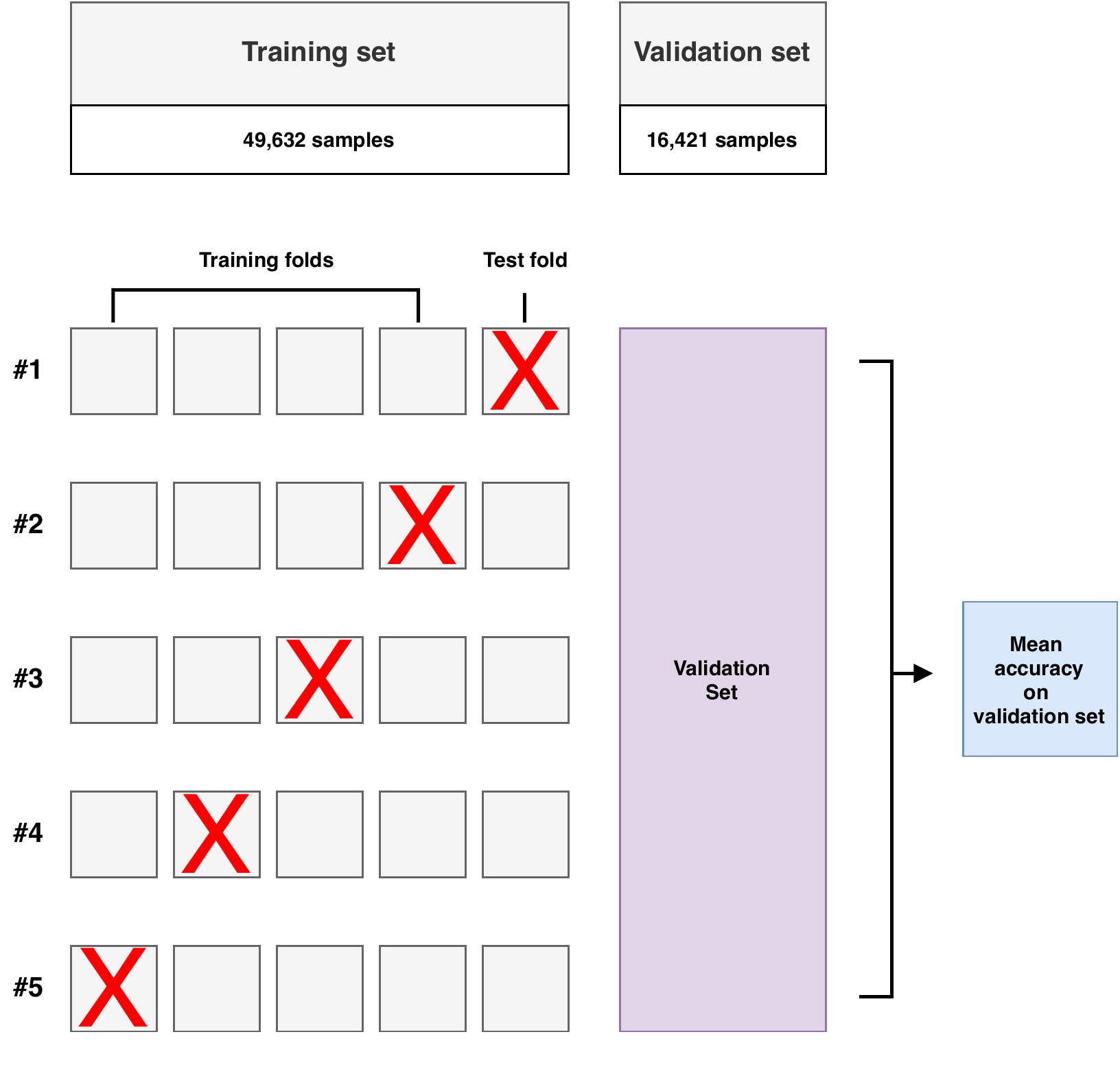}
    \caption{Visualisation of the model optimisation stage in the HICVS experiment with a large training set and 5-fold cross-validation}
    \label{fig:cross_validation_inductive__wv_larger_set}
\end{figure}

In the HICVS scenario, classifiers are trained on \emph{''Frame''} images and tested on \emph{''Comparison''} images, but in the model optimisation stage, a separate validation set is used, consisting of a subset of randomly, uniformly sampled examples from the \emph{''Comparison''} images.
The aim of this experiment was to determine and discuss the impact of applying GA in the model optimisation stage. The use of a validation set allows us to illustrate a scenario where GA can perform selection of features while maintaining model overfitting control. A discussion of this scenario will be presented in the Sec.~\ref{sec:discussion}.

In the HICVS scenario, all examples from tests scenes from all days are divided into two subsets - a test and a validation in a ratio of 80\% to 20\%. Similarly to the HIC scenario, pixels from test scene are not used as training examples. However, during model optimization stage, models are tested on the validation set. The test set consists of $65.676$ examples and the validation set consists of $16.421$ samples. In this scenario two values of the the training set size were considered to check the impact of extending the training set on classification accuracy. A variant with a smaller training set contains $6000$ examples ($250$ examples from each class, from four available images). In this variant of the experiment, models are trained using 10-fold cross-validation. In this case, the 9-fold samples form a training subset, and the model is tested on a validation set. The remaining fold is not involved in the validation process. Visualisation of the model optimisation process in the HICVS scenario with a small training set is presented in Fig.~\ref{fig:cross_validation_inductive__wv_smaller_set}. 

The second variant with a larger training set contains $49,632$ examples ($2068$ examples from each class, from four available images). The models are trained using 5-fold cross-validation, where 4 folds form a training subset, and the model is tested on validation set. As before, the remaining fold is not involved in the validation process.  Visualisation of the model optimisation process in the HICVS scenario with a large training set is presented in Fig.~\ref{fig:cross_validation_inductive__wv_larger_set}. 

After the optimisation process, the best model is tested on a test set that does not contain examples from the validation set.

\clearpage
\section{Results}
This section presents our results divided into the three scenarios corresponding to experiments described in Sec.~\ref{sec:experiments}. 

\subsection{The HTC scenario}
Results of HTC scenario are presented in Tab.~\ref{tab:results-htc}. We can see that the accuracy on the \emph{Comparsion} scene is much higher than in other experiments. Only the KNN classifier did not achieve an accuracy higher than $90\%$. We can also notice that the model based on the MLP classifier optimized with GS had the best result in each of the three days. In this scenario, the accuracies on \emph{Frame} images of all tested models were close to $100\%$ 

\ctable[
cap     = HTC results,
caption = Results of the HTC scenario for classification with GA and reference classifiers trained with grid search (GS). The highest result in each day is denoted with bold font.,
label   = tab:results-htc,
pos     = ht]
{llcccc}{\tnote{Denotes SVM with a linear kernel}\tnote[b]{Results for combined data from all days}}{\FL
Model&Classifier&\multicolumn{4}{c}{Accuracy/day}\NN
optimisation&&1&7&21&all\tmark[b]\ML
\multirow{5}{*}{GS}&SVC&$97.65\pm0.09$&$98.2\pm0.07$&$95.66\pm0.13$&$97.14\pm0.04$\NN
&LSVC\tmark[a]&$91.19\pm0.63$&$91.25\pm0.54$&$87.89\pm0.44$&$90.13\pm0.52$\NN
&$nu-$SVM&$97.02\pm0.1$&$97.87\pm0.12$&$95.27\pm0.15$&$96.66\pm0.07$\NN
&KNN&$87.96\pm0.17$&$90.3\pm0.2$&$85.22\pm0.21$&$87.65\pm0.12$\NN
&MLP&$\bm{98.94\pm0.10}$&$\bm{99.32\pm0.06}$&$\bm{98.31\pm0.12}$&$\bm{98.83\pm0.07}$\ML
GA&$nu-$SVM&$98.05\pm0.11$&$98.51\pm0.07$&$96.48\pm0.11$&$97.66\pm0.05$\ML
}
\subsection{The HIC scenario}
Results of the HIC scenario are presented in Tab.~\ref{tab:results-hic}. We can see that the classifier trained with GA outperforms reference methods only in the first day and even then the ranges of standard deviations overlap. For the remaining days the SVM with a linear kernel scored best. Interestingly, the best kernel chosen by GA optimisation was also the the linear kernel, and the number of bands has been reduced from 113 to 61. We also noticed that the training accuracy i.e. accuracy measured on the training set during model optimisation was close to 100\% for almost all models including the classifier trained with GA. 

\ctable[
cap     = HIC results,
caption = Results of the HIC scenario for classification with GA and reference classifiers trained with grid search (GS). The highest result in each day is denoted with bold font.,
label   = tab:results-hic,
pos     = ht]
{llcccc}{\tnote{Denotes SVM with a linear kernel}\tnote[b]{Results for combined data from all days}}{\FL
Model&Classifier&\multicolumn{4}{c}{Accuracy/day}\NN
optimisation&&1&7&21&all\tmark[b]\ML
\multirow{5}{*}{GS}&SVC&$70.42\pm0.94$&$62.42\pm1.09$&$60.4\pm0.63$&$65.06\pm0.83$\NN
&LSVC\tmark[a]&$72.89\pm0.33$&$\bm{67.79\pm0.64}$&$\bm{62.23\pm0.85}$&$\bm{68.08\pm0.52}$\NN
&$nu-$SVM&$69.69\pm0.42$&$58.73\pm0.55$&$56.93\pm0.33$&$62.66\pm0.39$\NN
&KNN&$66.29\pm0.3$&$58.74\pm0.13$&$55.11\pm0.19$&$60.66\pm0.06$\NN
&MLP&$67.64\pm0.49$&$62.01\pm0.94$&$59.69\pm0.52$&$63.57\pm0.60$\ML
GA&$nu-$SVM&$\bm{73.27\pm0.55}$&$63.42\pm0.75$&$60.54\pm0.48$&$66.54\pm0.45$\ML
}

\subsection{HIC scenario with a validation set}
Results of the HICVS scenario experiments are presented in Tab.~\ref{tab:results-hicvs}. In this scenario, accuracy of almost all classifiers improved compared to the HIC scenario (see Tab.~\ref{tab:results-hic}), but the GA-optimised classifier outperformed other methods. However, we also notice almost fourfold increase in standard deviation for GA optimised model. Once again, the linear kernel was the winning model for GA and 64 bands were selected. Similarly to the HIC scenario, the training accuracy i.e. accuracy measured on the training set during model optimisation was close to 100\% for almost all models including the classifier trained with GA. 

\ctable[
cap     = HICVS results,
caption = Results of the HICVS scenario for classification with GA and reference classifiers trained with grid search (GS). The highest result in each day is denoted with bold font.,
label   = tab:results-hicvs,
pos     = ht]
{llcccc}{\tnote{Denotes SVM with a linear kernel}\tnote[b]{Results for combined data from all days}}{\FL
Model&Classifier&\multicolumn{4}{c}{Accuracy/day}\NN
optimisation&&1&7&21&all\tmark[b]\ML
\multirow{5}{*}{GS}&SVC&$71.29\pm0.65$&$64.16\pm1.21$&$62.85\pm0.99$&$66.67\pm0.88$\NN
&LSVC\tmark[a]&$72.79\pm0.44$&$68.4\pm0.81$&$62.8\pm1.12$&$68.38\pm0.66$\NN
&$nu-$SVM&$71.75\pm0.95$&$63.86\pm1.69$&$63.67\pm1.24$&$67.04\pm1.16$\NN
&KNN&$67.59\pm0.47$&$60.36\pm0.14$&$55.9\pm0.38$&$61.88\pm0.1$\NN
&MLP&$67.49\pm0.44$&$62.14\pm1.02$&$59.68\pm0.83$&$63.54\pm0.66$\ML
GA&$nu-$SVM&$\bm{75.17\pm1.91}$&$\bm{73.42\pm2.11}$&$\bm{69.99\pm2.54}$&$\bm{73.02\pm2.14}$\ML
}

\section{Discussion}
\label{sec:discussion}
\subsection{The impact of preprocessing}
The preprocessing described in the Sec.~\ref{sec:extraction} is aimed to extract class features that are similar in all images. In order to illustrate the impact of the proposed preprocessing and data transformation on classification accuracy in HTC and HIC scenarios, we performed a simple experiment: We repeated the HIC scenario, i.e. we trained the $\nu-$SVM classifier obtained in the optimization process during the HIC scenario (including feature selection) with examples from all \emph{Frame} images. However, we omitted the the step III `feature extraction' of the procedure described in Sec.~\ref{sec:experiments} i.e. the classifier processed normalised spectra. The training set size was 6000 examples (250 examples from each class, from four available images). The accuracy on the combined \emph{Comparison} images was $acc'_{\text{Comp.}}=54.02\pm0.21$ which is lower than the corresponding value in the Tab.~\ref{tab:results-hic} i.e. $acc_{\text{Comp}}=66.54\pm0.45$. At the same time, the accuracy on the remaining pixels of \emph{Frame} images was $acc'_{\text{Frame}}=99.61\pm0.05$ which is similar to results of HTC experiments.

We can see that in the HTC scenario when training and testing examples come from the same scene, the classifier can  model classes and reach high classification accuracy even without preprocessing. However, the proposed preprocessing improves the accuracy in the HIC scenario, when training and test data are more different.

\subsection{Model optimisation with GA in hyperspectral classification}
Reference work on hyperspectral GA-based classification described in Sec.~\ref{sec:SOA} present their advantages such as the reduction in data dimensionality through band selection, their resistance to overfiting or their consistently higher accuracy than for the reference model selected with GS~\cite{sukawattanavijit2017ga}. 
However, most of the works consider only the HTC scenario, use similar, airplane or satellite-based images and sometimes compare the method with a model trained with preset parameters~\cite{zhuo2008genetic}. Therefore, to better assess the capability of GA-based model selection, we compared GA and GS in two scenarios that differ in regards to the complexity of the classification problem. 

Our results show that in the HTC scenario, both model optimisation techniques resulted in comparable, highly accurate models. We noticed that the accuracy measured on the training set during the process of model optimisation, was very similar to the final accuracy on the test set.
It seems that for training and test sets created by randomly, uniformly sampling a hyperspectral image, spectra in both sets are similar enough that GA and GS-based model are similar in regards to their accuracy and the major advantage of GA in this scenario is the band selection, which more than halved the number of features in our experiments. 

Compared to the HTC, the HIC scenario proved to be significantly more challenging. Accuracy values in Tab.~\ref{tab:results-hic} are lower and it seems that GA-trained classifier was only slightly better than GS for images captured in the first day and scored second for  test images captured in other days (although, the number of features was once again halved).  
In the HIC scenario, training and test data come from images that differ in regards to lightning conditions and spectral mixtures of imaged classes and the image background. 
We hypothesise, that despite the fact that both images contain the same, precisely applied and clearly visible substances, differences between the training and the test set are so significant, that the selected model is overfitted. This is supported by the fact that similarlty to the HTC scenario, the accuracy measured on the training set during the process of model optimisation was very high in the HIC. While GA allows local maxima to be avoided during model optimization, when all training data is noisy in the same way, there is no global maximum that GA could find. This hypothesis is further supported by the higher accuracy of the method on the first day images. Images acquired in the first day are more similar since aging has significant impact on specta e.g. the `blood' class spectrum changes significantly~\cite{majda2018hyperspectral} due to hemoglobin oxidation.

In order to better explore the capabilities of GA in HSI model optimisation, we proposed one more experiment: the HICVS scenario described in detail in Sec.~\ref{sec:exp:HICVS}. In HICVS, the classifier is trained on a similar training set as in the HIC scenario, but during the model optimisation stage, the optimisation algorithm has an access to examples in the validation set that are similar to test data. We expect that in this situation GA should gain an observable advantage over GS: since the algorithm can now control model overfitting through every epoch it should be able to create a better generalizing classifier. Results in Tab.~\ref{tab:results-hicvs} confirm this hypothesis: while results of GS also improved, the improvement for GA is higher and it scored first for all images.

Referring to our initial hypothesis introduced in Sec.~\ref{sec:introduction}, that GA allows to obtain more accurate hyperspectral classifiers than GS: in our opinion presented results support this hypothesis, provided that certain assumptions related to the nature of the processed hyperspectral images are met. First: for a uniform data set as e.g. in the HTC scenario, when the training set is sufficient and uniformly sampled, both model optimisation methods can result in highly accurate, comparable classifiers. However, when spectra become noisy which results in differences between the training and test sets, GA can outperform GS and avoid model overfitting, provided that a subset of examples similar to test data are available during model optimisation. However, when the noise between training and test data becomes too big, the advantage of GA over GS in terms of accuracy seems not significant. However, compared to GS, in all scenarios, GA can produce similar or more accurate classifiers while at the same time significantly reducing the dimensionality of the data through band selection.

\section{Conclusion and future works}
We compared GA-based model selection with classic approach based on grid search in three different hyperspectral classification scenarios. In the \emph{Hyperspectral Transductive Classification} (HTC) scenario, the training and test data are taken from a single image, so they are similar. In the \emph{Hyperspectral Inductive Classification} (HIC) scenario, the training and test data come from different images. The third scenario i.e. the \emph{Hyperspectral Inductive Classification with a Validation Set} (HICVS) was created on the basis of the HIC scenario, in which the model selection algorithm has access to examples similar to those in the test set. Our results show that for noisy data, as in HIC, the advantage of GA over GS in terms of accuracy is not significant and that in order to achieve this advantage, GA must have examples representative of the test set at the model selection stage as e.g. in the HICVS scenario. However, in all tested scenarios GA was able to generate similarly or more accurate models than GS while significantly reducing the dimensionality of data through band selection. 

We plan to apply our approach to different models, in particular recurrent neural networks, deep neural networks and ensemble learning. We would also like to test different feature extraction methods dedicated to GA-based classification of hyperspectral images, especially in the HIC scenarios.

\section{Acknowledgement}
K.K. acknowledges funding from the European Union through the European Social Fund (grant POWR.03.02.00-00-I029).

\printbibliography

\end{document}